\documentclass[a4paper,twoside]{article}

\usepackage{epsfig}
\usepackage{subcaption}
\usepackage{calc}
\usepackage{amssymb}
\usepackage{amstext}
\usepackage{amsmath}
\usepackage{amsthm}
\usepackage{multicol}
\usepackage{pslatex}
\usepackage{apalike}

\usepackage{verbatim}
\usepackage{courier}
\usepackage{booktabs} 
\usepackage{arydshln}
\usepackage{multirow}
\usepackage{makecell}
\usepackage[highlight]{mavilab}
\usepackage[compact]{titlesec}
\usepackage{soul}

\usepackage{fancyhdr}
\usepackage[pagebackref=true,breaklinks=true,letterpaper=true,colorlinks,bookmarks=false]{hyperref}
\usepackage{SCITEPRESS}

\begin{document}

\title{Describing image focused in cognitive and visual details for visually impaired people: An approach to generating inclusive paragraphs}

\author{
\authorname{
Daniel L. Fernandes\sup{1}\orcid{0000-0002-6548-294X}, 
Marcos H. F. Ribeiro\sup{1}\orcid{0000-0003-4481-5781},  
Fabio R. Cerqueira\sup{1}\sup{2}\orcid{0000-0003-1325-2592},
and 
Michel M. Silva\sup{1}\orcid{0000-0002-2499-9619} 
}
\affiliation{\sup{1}Department of Informatics, Universidade Federal de Viçosa - UFV, Viçosa, Brazil}
\affiliation{\sup{2}Department of Production Engineering, Universidade Federal Fluminense - UFF, Petrópolis, Brazil}
\email{\{daniel.louzada, marcosh.ribeiro, michel.m.silva\}@ufv.br, frcerqueira@id.uff.br}
}


\abstract{Several services for people with visual disabilities have emerged recently due to achievements in Assistive Technologies and Artificial Intelligence areas. Despite the growth in assistive systems availability, there is a lack of services that support specific tasks, such as understanding the image context presented in online content, \eg, webinars. Image captioning techniques and their variants are limited as Assistive Technologies as they do not match the needs of visually impaired people when generating specific descriptions. We propose an approach for generating context of webinar images combining a dense captioning technique with a set of filters, to fit the captions in our domain, and a language model for the abstractive summary task. The results demonstrated that we can produce descriptions with higher interpretability and focused on the relevant information for that group of people by combining image analysis methods and neural language models.}

\onecolumn \maketitle \normalsize \setcounter{footnote}{0} \vfill

\thispagestyle{fancy}
\fancyhf{}
\chead{In Proceedings of the 17th International Joint Conference on Computer Vision, \\ Imaging and Computer Graphics Theory and Applications (VISAPP) 2022 \\ The final publication is available at: \href{https://doi.org/10.5220/0010845700003124}{doi.org/10.5220/0010845700003124}.}
\setlength{\headsep}{0.6 in}

\section{\uppercase{Introduction}}
\label{sec:introduction}

Recently, we have witnessed a boost in several services for visually impaired people due to achievements in Assistive Technology and Artificial Intelligence~\cite{alhichri2019helping}. However, the problems faced by the impaired people are literally everywhere, from arduous and complex challenges, such as going to groceries, to daily tasks like recognizing the context of TV news, online videos or webinars. In the last couple of years, online content and videoconferencing tools have been widely used to overcome the restrictions imposed by the social distance of the COVID-19 pandemic~\cite{wanga2020social}. Nonetheless, there is a lack of tools that allow visually impaired people to understand the overall context of such content~\cite{gurari2020captioning,simons2020hope}. 

One of the services applied as an Assistive Technology is the Image Captioning techniques~\cite{gurari2020captioning}. 
Despite the remarkable results for the automatically generated image caption, these techniques are limited when applied to extract the image context. 
Even with recent advances to improve the quality of the captions, at the best, they compress all the visual elements of an image into a single sentence, resulting in a simple generic caption~\cite{ng2020understanding}. Since the goal of Image Captioning is to create a high-level description for the image content, the cognitive and visual details needed by visually impaired people are disregarded~\cite{dognin2020image}.

To address the drawback of Image Captioning about enclosing all visual details in a single sentence, Dense Captioning creates a descriptive sentence for each meaningful image region~\cite{densecap}. Nonetheless, such techniques fail to synthesize information into coherently structured sentences due to the overload caused by multiple disconnected sentences to describe the whole image~\cite{krause2016paragraphs}. Thus, Dense Captioning is not suitable to the task of extracting image context for impaired people since it describes every visual detail of the image in a unstructured manner, not providing cognitive information.

Image Paragraph techniques address the demand for generating connected and descriptive captions~\cite{krause2016paragraphs}. The goal of these techniques is to generate long, coherent and informative descriptions about the whole visual content of an image~\cite{chatterjee2018diverse}. Although being capable of generating connected and informative sentences, the usage of these techniques is limited when creating image context for visually impaired people, because these techniques have generic application and result in a long description to tell a story about the image and all its elements. By creating a long and dense paragraph that describes the whole image, the needs of those with visual impairment are not matched.     

The popularity of the social inclusion trend is noticeable in social media, \eg, the Brazilian project \#ForBlindToSee (from the Portuguese, \#PraCegoVer) has reached trend topics in Twitter. This project provides a manual description for an image to be reproduced by audio tools for the visually impaired, and also encourages people during videoconferences to perform an oral description of the main characteristics of their appearance and the environment where they are located. The description helps the visually impaired audience to better understand the context of the reality of the person who is presenting. Although this is a promising inclusion initiative, once the oral descriptions are dependent on sighted people, it may not be provided or may even be poorly descriptive.

Motivated to mitigate the accessibility barriers for vision-impaired people, we combine the advantages of the Dense Captioning techniques, task-specific Machine Learning methods and language models pretrained on massive data. Our method consists in a pipeline of computational techniques to automatically generate suitable descriptions for this specific audience about the relevant speaker's characteristics (\eg, physiognomic features, facial expressions) and their surroundings~\cite{brasil2016,lewis2018}.

\section{\uppercase{Related Work}}
\label{sec:related}

Works on image description and text summarization have been extensively studied. We provide an overview classifying them in the following topics.

\begin{figure*}[h]
  \centering
   \includegraphics[width=1.0\textwidth]{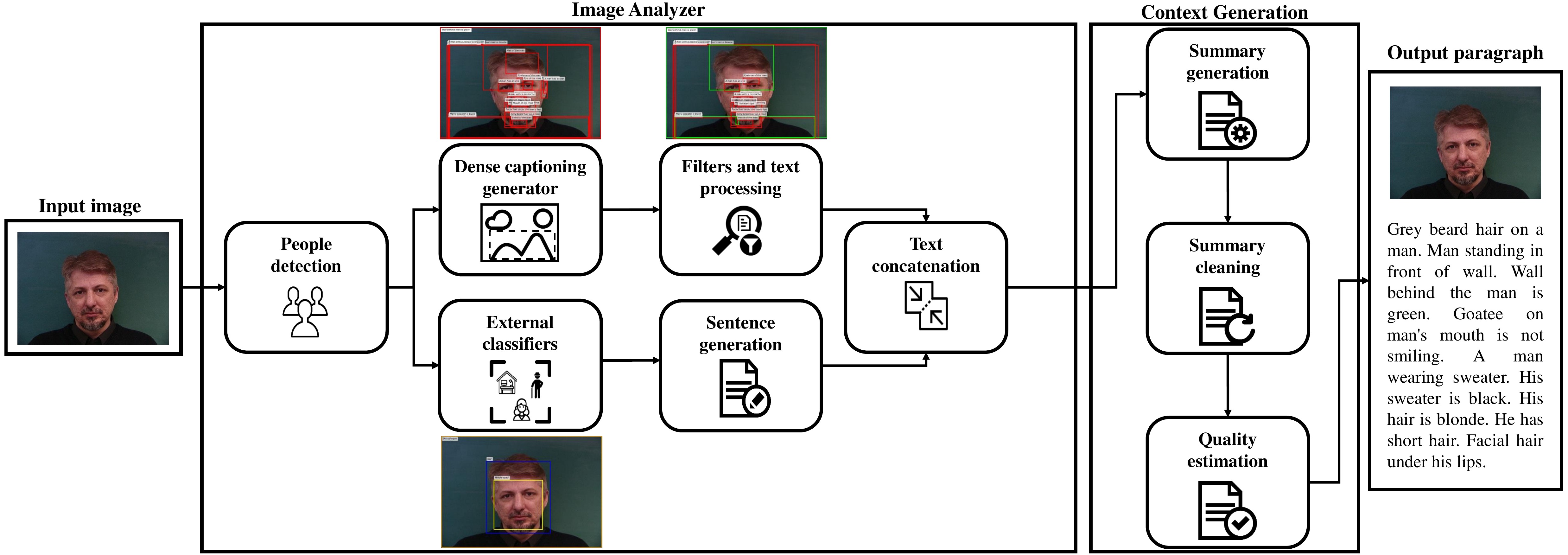}
   \caption{Given a webinar image, its relevant information along with speaker attributes from pretrained models are extracted into textual descriptions. After a filter processing to fit the person domain, all descriptions are  aggregated into a single text. We summarize the generated single text, and after a cleaning and filtering process, the best quality paragraph is selected.}
  \label{fig:fig1}
\end{figure*}

\paragraph{Image Captioning.}

Image Captioning aims the generation of descriptions for images and has attracted the interest of researchers, connecting Computer Vision and Natural Language Processing. 

Several frameworks have been proposed for the Image Captioning task based on deep encoder-decoder architecture, in which an input image is encoded into an embedding and subsequently decoded into a descriptive text sequence~\cite{ng2020understanding,vinyals2015show}. Attention Mechanisms and their variations were implemented to incorporate visual context by selectively focusing on the specific part of the image~\cite{xu2015show} and to decide when to activate visual attention by means of adaptive attention and visual sentinels~\cite{lu2017knowing}.

Most recently, a novel framework for generating coherent stories from a sequence of input images was proposed by modulating the context vectors to capture temporal relationship on the input image sequence using bidirectional Long Short-Term Memory (bi-LSTM)~\cite{malakan2021contextualise}. To maintain the image specific relevance and context, image features and context vectors from the bi-LSTM are projected into a latent space and submitted to an Attention Mechanism to learn the spatio-temporal relationships among image and context. The encoder output is then modulated with the input word embedding to capture the interaction between the inputs and their context using \textit{Mogrifier}-LSTM that generates relevant and contextual descriptions of the images while maintaining the overall story context. 

Herdade~\etal proposed a Transformer architecture with an encoder block to incorporate information about the spatial relationships between input objects detected through geometric attention~\cite{Herdade2019ImageCT}. Liu~\etal also addressed the captioning problem using Transformers by replacing the CNN-based encoder of the network by a Transformer encoder, reducing the convolution operations~\cite{Liu2021CPTRFT}.

Different approaches have been proposed to improve the discriminative capacity of the generated captions and attenuate the restrictions presented in previously proposed  methods. Despite the efforts, most models still produce generic and similar captions~\cite{ng2020understanding}.

\paragraph{Dense Captioning.}

Since region-based descriptions tend to be more detailed than global descriptions, the Dense Captioning task aims at generalizing the tasks of Object Detection and Image Captioning into a joint task by simultaneously locating and describing salient regions of an image~\cite{densecap}. 
Johnson~\etal proposed a Fully Convolutional Localization Network architecture, which supports end-to-end training and efficient test-time performance, composed of a CNN to process the input image and a dense localization layer to predict a set of regions of interest in the image. The descriptions for each region is created by a LSTM with natural language previously processed by a fully connected network~\cite{densecap}. 
Additional approaches improve Dense Captioning by using joint inference and context fusion~\cite{yang2017dense}, and the visual information about the target region and multi-scale contextual cues~\cite{yin2019context}.

\paragraph{Image Paragraphs.}

Image Paragraphs is a captioning task that combines the strengths of Image Captioning and Dense Captioning, generating long, structured and coherent paragraphs that richly describe the images~\cite{krause2016paragraphs}. This approach was motivated by the lack of detail described in a single high-level sentence of the Image Captioning techniques and the absence of cohesion when describing whole image due to the large amount of short independent captions returned by Dense Captioning techniques.

A pioneering approach proposed a two-level hierarchical RNN to decompose paragraphs into sentences after separating the image into regions of interest and aggregating the features of these regions in semantically rich representation~\cite{krause2016paragraphs}. This approach was extended with an Attention Mechanism and a GAN that enables the coherence between sentences by focusing on dynamic salient regions~\cite{liang2017recurrent}. An additional study uses coherence vectors and Variational Auto-Encoder to increase sentence consistency and paragraph diversity~\cite{chatterjee2018diverse}. 

Our approach diverges from the Image Paragraph. Instead of unconstrained describing the whole image, we focus on the relevant speaker's attributes to provide a better understanding to the visually impaired.

\paragraph{Abstractive Text Summarization.}

This task aims to rewrite a text into a shorter version, keeping the meaning of the original content.
Recent work on Abstractive Text Summarization was applied in different problems, such as highlighting news~\cite{zhu2021make}. 
Transformers pre-trained in massive datasets achieved remarkable performance on many natural language processing tasks~\cite{raffel2019exploring}. In Abstractive Text Summarization, we highlight the state-of-the-art pre-trained models BART, T5, and PEGASUS, with remarkable performance in both manipulated use of lead bias~\cite{zhu2021make} and zero or few-shot learning settings~\cite{goodwin2020flight}.

\section{\uppercase{Methodology}}

\label{sec:method}

Our approach takes an image as input and generates a paragraph containing the image context for visually impaired people, regarding context-specific constraints. As depicted in Fig.~\ref{fig:fig1}, the method consists of two phases: Image Analyzer and Context Generation.

\subsection{Image Analyzer}
\label{subsubsec:ImageAnalyzer}

We generate dense descriptions from the image, filter the descriptions, extract high-level information from the image, and aggregate all information into a single text, as presented in the following pipeline.

\paragraph{People detection.}
\label{subsubsec:people}

Our aim is to create an image context by describing the relevant characteristics about the speaker and their surroundings in the context of webinars. Therefore, as a first step, we apply a people detector and count the number of people $P$ in the image.  
If ${P = 0}$, the process is stopped.

\paragraph{Dense captioning generator.}

Next, we use an automatic Dense Captioning approach to produce independent and short sentence description for each meaningful region of the image. Every created description also has a confidence value and a bounding box.

\paragraph{Filters and text processing.}
\label{subsubsec:filter}

The Dense Captioning produces a large number of captions per image, which could result in similar descriptions or even out of context. In this step, we filter out those captions. 
Initially, we convert all captions to lower case, remove punctuation and duplicated blank spaces, and apply a tokenization method in the words.
Then, to reduce the amount of similar dense captions, we create a high-dimensional vector representation for each sentence using word embedding and calculate the cosine similarity between all sentence representations. Dense captions with cosine similarity greater than threshold $T_{text\_sim}$ are discarded. 

Next, we remove descriptions out of our context, \ie, captions that are not associated with the speaker. Linguistic annotation attributes are verified for each token in the sentences. If no tokens related to nouns, pronouns or nominal subjects are found, the sentence is discarded. Then, a double check is performed over the sentences that were kept after this second filtering. This is done using WordNet cognitive synonym sets (synsets)~\cite{miller1995wordnet}, verifying whether the tokens are associated to the concept person, being one of its synonyms or hyponym~\cite{krishnavisualgenome}.

It is important to filter out short sentences since they are more likely to have obvious details and to be less informative for visually impaired people~\cite{brasil2016,lewis2018}, for example, ``a man has two eyes'', ``the man's lips'', \etc. We remove all sentences shorter than the median length of the image captions.

On the set of kept sentences, we standardize the subject/person in the captions using the most frequent one, aiming to achieve better results in the text generator.
At the end of this step, we have a set of semantically diverse captions, related to the concept of person and the surroundings. Due to the filtering process, the kept sentences tend to be long enough to not contain obvious or useless information.

\paragraph{External classifiers.}

To add information regarding relevant characteristics for vision-impaired people, complementing the Dense Captioning results, we apply people-focused learning models, such as age detection and emotion recognition, and a scene classification model.

Due to the challenge of predicting the correct age of a person, we aggregate the returned values by the age group proposed by the World Health Organization~\cite{ahmad2001age}, \ie, Child, Young, Adult, Middle-aged, and Elderly.
Regarding emotion recognition and scene classification models, we use their outputs in cases where the model confidence is greater than a threshold $T_{model\_confidence}$. To avoid inconsistencies in the output, age and emotion models are only applied if a single person is detected in the image.

\paragraph{Sentence generation and text concatenation.}

From the output produced by the external classifiers, coherent sentences are created to include information about age group, emotion, and scene in the set of filtered descriptions for the input image. The generated sentences follow the default structure: “there is a/an $<$AGE$>$ $<$NOUN$>$", “there is a/an $<$NOUN$>$ who is $<$EMOTION$>$", and “there is a/an $<$NOUN$>$ in the $<$SCENE$>$", where AGE, EMOTION and SCENE are the output of the learning methods, and the $<$NOUN$>$ is the frequent person-related noun used to standardize the descriptions. Example of generated sentences, “there is a middle-aged woman", “there is a man who is sad", and “there is a boy in the office". These sentences are concatenated with the set of previously filtered captions into a single text.

\subsection{Context Generation}
\label{subsubsec:ParagraphGeneration}

In this phase, a neural linguist model is fed with the output of the first phase and generates coherent and connected summary, which goes through a new cleaning process to create a quality image context.

\paragraph{Summary generation.}

To create a human-like sentence, \ie, a coherent and semantically connected structure, we apply a neural language model to produce a summary from the concatenated descriptions resulting from the previous phase. Five distinct summaries are generated by the neural language model.

\paragraph{Summary cleaning.}
\label{subsubsec:summary_cleaning}

One important step is to verify if the summary produced by the neural language model achieves high similarity with the input text provided by the Image Analyzer phase. We assure the similarity between the language model output and its input, by filtering out phrases inside the summary when cosine similarity is less than threshold $\alpha$. A second threshold $\beta$ is used as an upper limit to the similarity values between pairs of sentences inside the summary to remove duplicated sentences. In pairs of sentences in which the similarity is greater than $\beta$, one of them is removed to decrease redundancy. As a result, after the application of the two threshold-based filters, we have summaries that are related to the input with low probability of redundancy.

\paragraph{Quality estimation.} 

After generating and cleaning the summaries, it is necessary to select one summary returned by the neural language model. We model the selection process to address the needs of vision-impaired people in understanding an image context by estimating the quality of paragraphs.

The most informative paragraphs usually present linguistic characteristics, such as, multiple sentences connected by conjunctions, use of complex linguistic phenomena (\eg, co-references), and have a higher frequency of verbs and pronouns~\cite{krause2016paragraphs}. Aiming to return the most informative summary, for each of the five filtered summaries, we calculate the frequency of the aforementioned linguistic characteristics. Finally, the output of our proposed approach is the summary with the higher frequency of these characteristics.

\section{\uppercase{Experiments}}
\label{sec:experiments}

Since there is no labeled dataset and evaluation metrics for the task of image context generation for visually impaired people, we adapted an existing dataset and used general purpose metrics, as described in this section.

\paragraph{Dataset.}

We used the Stanford Image-Paragraph (SIP) dataset~\cite{krause2016paragraphs}, widely applied for visual Image Paragraph task. SIP is a subset of ${19{,}551}$ images from the Visual Genome (VG) dataset~\cite{krishnavisualgenome}, with a single human-made paragraph for each image. VG was used to access human annotations of SIP images. 

We selected only the images related to our domain of interest, by analyzing their VG dense human annotations. Only images with at least one dense caption concerning to person were kept. To know whether a dense caption is associated with a person, we used WordNet synsets. To perform a double check about the presence and relevance of people in the image, we used a people detector, and filter out all images in which no person were detected or the ones that do not have a person bounding box with area greater than ${50\%}$ of the image area. Our filtered SIP dataset consists of ${2{,}147}$ images. Since the human annotations of SIP could contain sentences beyond our context, we filtered out sentences that were not associated with a person by means of linguistic feature analysis.

\paragraph{Implementation details.}

For people detection and dense captioning tasks, we used YOLOv3~\cite{Redmon2018YOLOv3} with minimum probability of ${0.6}$, and DenseCap~\cite{densecap}, respectively. To reduce the amount of similar dense captions, we adopted $T_{text\_sim} = 0.95$. For age inference, emotion recognition, and scene classification tasks, we used, respectively, Deep EXpectation~\cite{rothe2015dex},  Emotion-detection~\cite{goodfellow2013challenges}, and  PlacesCNN~\cite{zhou2017places}. For emotion and scene models, we used $T_{model\_confidence}=0.6$.  

We used a T5 model~\cite{raffel2019exploring} (Huggingface \texttt{T5-base} implementation) fine-tuned on the News Summary dataset for abstractive summarization task and with beam search widths in the range of 2-6. To keep sentences relevant and unique, we defined $\alpha$ and $\beta$ equal to ${0.7}$ and ${0.5}$, respectively.

\paragraph{Competitors.}

We compared our approach with two competitors, Concat, that creates a paragraph by concatenating all DenseCap outputs, and Concat-Filter, that concatenates only the DenseCap outputs associated with a person, as described in Section~\ref{subsubsec:filter}. The proposed method is mentioned as Ours hereinafter, and a variant is referred as Ours-GT. This variant uses the human-annotated dense captions of the VG database instead of using DenseCap for creating the dense captions for the images, and it is used as an upper-bound comparison for our method.

\begin{table}[t]
\caption{Comparison between methods using standard metrics to measure similarity between texts. All values are reported as percentage. B, Mt and Cr metrics stand for BLEU, METEOR and CIDEr, respectively. Best values in bold.}
\label{tab:tab1}
\small
\setlength{\tabcolsep}{5.3pt}
\begin{tabular}{ l r r r r r r }
  \toprule
  \thead{\textbf{Method}} & \thead{B-1} & \thead{B-2} & \thead{B-3} & \thead{B-4} & \thead{Mt} & \thead{Cr}\\
  \cmidrule(rl){2-5} \cmidrule(rl){6-6} \cmidrule(rl){7-7}
  Concat & 6.4 & 3.9 & 2.3 & 1.3 & 11.4 & 0.0\\
  Concat-Filter & \textbf{29.8} & \textbf{17.5} & \textbf{9.9} & \textbf{5.5} & \textbf{16.4} & \textbf{9.6}\\
  Ours & 16.9 & 9.6 & 5.3 & 3.0 & 10.9 & 9.1\\
  \cdashline{2-7}
  \textit{Ours-GT} & \textit{22.3} & \textit{12.1} & \textit{6.6} & \textit{3.7} & \textit{12.5} & \textit{15.3}\\
  \bottomrule
\end{tabular}
\end{table}

\paragraph{Evaluation metrics.}

The performance of the methods was measured through the metrics: BLEU-\{1, 2, 3, 4\}, METEOR and CIDEr, commonly used on paragraph generation and image captioning tasks.

\begin{table}[b]
\caption{Language statistics about the average and standard deviation in the number of characters, words, and sentences in the paragraphs generated by each method. }
\label{tab:tab2} 
\small
\setlength{\tabcolsep}{6.2pt}
\begin{tabular}{ l r r r }
  \toprule
  \thead{\textbf{Method}} & 
  \thead{Characters} &
  \thead{Words} &
  \thead{Sentences} \\
  \cmidrule(rl){2-4}
  Concat & 2,108 $\pm$ 311 & 428 $\pm$ 58 & 89 $\pm$ 12 \\
  Concat-Filter & 294 $\pm$ 116 & 62 $\pm$ 23 & 12 $\pm$ 5\\
  Ours & 109 $\pm$ 42 & 22 $\pm$ 8 & 3 $\pm$ 1\\
  Ours-GT & 139 $\pm$ 54 & 26 $\pm$ 10 & 3 $\pm$ 1\\
  \cmidrule(rl){2-4}
  \textit{Humans} & \textit{225} $\pm$ \textit{110} & \textit{45} $\pm$ \textit{22} & \textit{4} $\pm$ \textit{2}\\
  \bottomrule
\end{tabular}
\end{table}
\begin{table*}[h]
\caption{Comparing Ours with the filtered Concat-Filter to match the number of sentences written by humans. Lines bellow the dashed show the metrics when the filtering is smoothed. All values are reported as percentage, and best ones are in bold. 
}
\label{tab:tab3} \centering
\small

\begin{tabular}{ l r r r r r r }
  \toprule
  \thead{\textbf{Method}} & \thead{BLEU-1} & \thead{BLEU-2} & \thead{BLEU-3} & \thead{BLEU-4} & \thead{METEOR} & \thead{CIDEr}\\
  \cmidrule(rl){2-5} \cmidrule(rl){6-6} \cmidrule(rl){7-7}
  Concat-Filter-25\% & 7.4 & 4.0 & 2.1 & 1.1 & 8.2 & 4.7\\
  Ours & \textbf{16.9} & \textbf{9.6} & \textbf{5.3} & \textbf{3.0} & \textbf{10.9} & \textbf{9.1}\\
  \cdashline{2-7}
  Concat-Filter-30\% & 11.3 & 6.1 & 3.2 & 1.7 & 9.2 & 6.2\\
  Concat-Filter-35\% & 16.0 & 8.7 & 4.6 & 2.4 & 10.4 & 8.1\\
  Concat-Filter-40\% & 19.4 & 10.6 & 5.6 & 3.0 & 11.0 & 9.0\\
  Concat-Filter-45\% & 22.6 & 12.4 & 6.6 & 3.6 & 11.7 & 9.9\\
  \bottomrule
\end{tabular}
\end{table*}
\begin{table}[h] 
\caption{Statistics of paragraph linguistic features for Nouns (N), Verbs (V), Adjectives (A), Pronouns (P), Coordinating Conjunction (CC), and Vocabulary Size (VS). All values are reported as percentage, except for VS. Values in bold are the closest to the human values.}
\label{tab:tab4}
\small
\setlength{\tabcolsep}{5.5pt}
\begin{tabular}{ l r r r r r r }
  \toprule
  \thead{\textbf{Method}} & \thead{N} & \thead{V} & \thead{A} & \thead{P} & \thead{CC} & \thead{VS}\\
  \cmidrule(rl){2-7}
  Concat & \textbf{30.3} & 3.3 & 14.3 & 0.0 & \textbf{1.9} & \textbf{876}\\
  Concat-Filter & 32.6 & 8.6 & 8.4 & 0.1 & 0.5 & 418\\
  Ours & \textbf{30.3} & \textbf{10.6} & \textbf{9.9} & \textbf{1.6} & 1.5 & 642\\
  \cdashline{2-7}
  \textit{Humans} & \textit{25.4} & \textit{10.0} & \textit{9.8} & \textit{5.9} & \textit{3.1} & \textit{3,623} \\
  \bottomrule
\end{tabular}
\end{table}

\section{\uppercase{Results and Discussion}}
\label{sec:results}

We start our experimental evaluation analyzing Tab.~\ref{tab:tab1}, which shows the values of the metrics for each method. The Concat approach presented the worst performance due to the excess of independent produced captions, demonstrating its inability to produce coherent sentences~\cite{krause2016paragraphs}. 
When comparing with Concat, Ours achieved better performance in most metrics. 
The descriptions generated by the Concat-Filter method produced higher scores when compared with our approach. However, it is worth to note that Ours and Concat-Filter presented close values considering the CIDEr metric. Unlike the metrics BLEU~\cite{papineni2002bleu} and METEOR~\cite{banerjee2005meteor} that are intended for the Machine Translation task, CIDEr was specifically designed to evaluate image captioning methods based on human consensus~\cite{vedantam2015cider}.
Furthermore, the CIDEr value of the Concat-Filter is smaller than the value of Ours-GT, which demonstrates the potential of our method in the upper-bound scenario. 

Tab.~\ref{tab:tab2} presents language statistics, demonstrating that Concat-Filter generates medium-length paragraphs, considering the number of characters, and average number of words closer to paragraphs described by humans. However, its paragraphs have almost three times the number of sentences of human paragraphs. This wide range of information contained in the output explains the higher values achieved in BLEU-\{1,2,3,4\} and METEOR, since these metrics are directly related to the amount of words present in the sentences.

To further demonstrate that the good results achieved by Concat-Filter was due to the greater number of words, we randomly removed the sentences from the paragraphs of Concat-Filter. The average number of sentences were closer to paragraphs described by human when ${75\%}$ of the sentences were removed. Then, we ran the metrics again and the values for Concat-Filter dropped substantially, as demonstrated in Tab.~\ref{tab:tab3}. 
Considering ${25\%}$ of the sentences, Ours overcome the results in all metrics.
Concat-Filter only achieves comparable performance with Ours when it is kept at least ${40\%}$ of the total paragraph sentences. This is a higher amount compared to the average of sentences present in paragraphs created by humans and our approach (Tab.~\ref{tab:tab2}).

We can see in Tab.~\ref{tab:tab4} that Ours achieved results closer to the human annotated output for most of the linguistic features. The results presented in this section demonstrate that paragraphs generated by our approach are more suitable for people with visual impairments, since their resulting linguistic features are more similar to the ones described by humans, and mainly, the final output is shorter. As discussed and presented in Section~\ref{sec:qualitative_analysis}, the length is crucial to better understand the image context.

\subsection{Qualitative Analysis}
\label{sec:qualitative_analysis}

\begin{figure*}[!h]
  \centering
   {\epsfig{file = 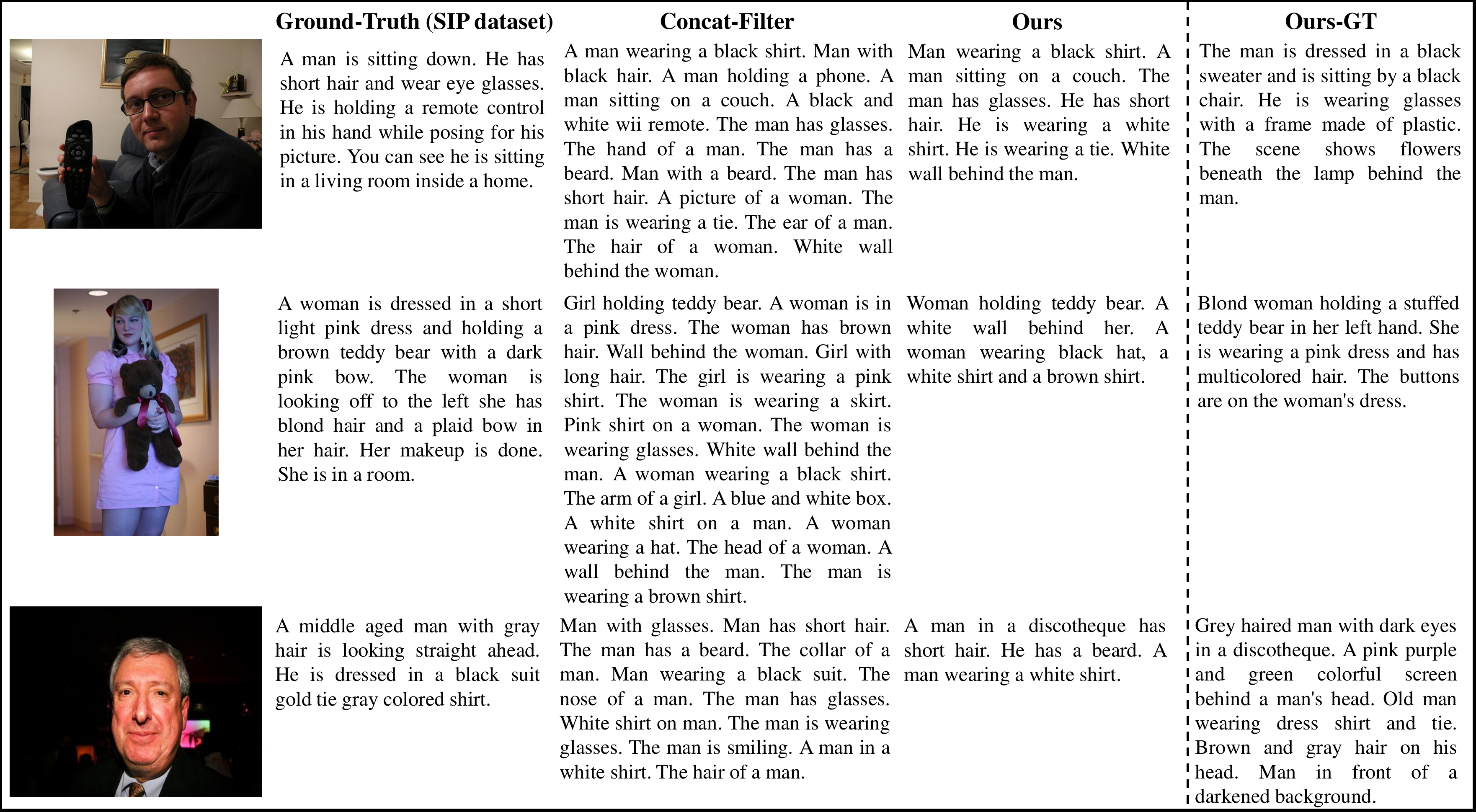, width = \textwidth}}
  \caption{Examples of paragraphs generated by the compared methods, except for Concat.}
  \label{fig:fig2}
 \end{figure*}

Fig.~\ref{fig:fig2} depicts input images and their context paragraphs generated by the methods compared in this study. Concat-Filter returned text descriptions with simple structure, containing numerous and obvious details in general. These observations corroborate with the data presented in Tab.~\ref{tab:tab4}, which show the low amount of pronouns and coordinating conjunctions.
In contrast, even using a language model that is not specific to the Image Paragraph task, Ours and Ours-GT generate better paragraphs in terms of linguistic elements, using commas and coordinating conjunctions, which smooth phrase transitions.

  \begin{figure*}[!h]
  \centering
   {\epsfig{file = 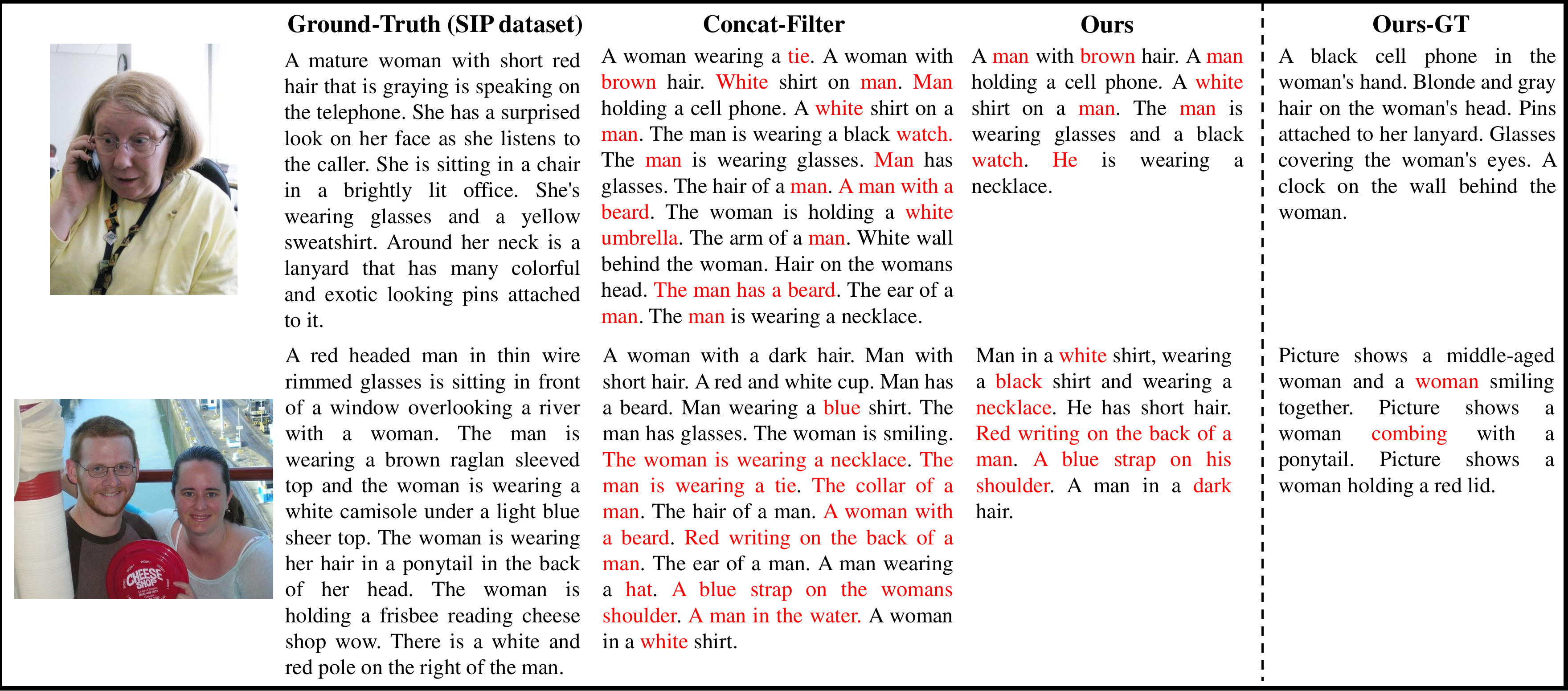, width = \textwidth}}
  \caption{Examples of paragraphs generated with different failure cases. Sentences and words in red are flawed cases.}
  \label{fig:fig3}
 \end{figure*}

In Fig.~\ref{fig:fig2} column 3, we can note some flaws in the generated context paragraphs. The main reason is the error propagated by the models used in the pipeline, including DenseCap. 
Despite that, the results are still intelligible and contribute to the image interpretability for visual-impaired people. Due to the flexibility of our approach, errors can be reduced by replacing models by similar ones with higher accuracy. 
One example is Ours-GT, in which DenseCap was replaced by human annotations and the results are better in all metrics and qualitative analysis, demonstrating the potential of our method.

Fig.~\ref{fig:fig3} illustrates failure cases generated by the methods. In the first line, a woman is present in the image and the output of Ours described her as a man, which can be justified since the noun man is more frequent in the DenseCap output than woman, as can be seen in the Concat-Filter column. This mistake was not made by Ours-GT, since this approach does not use DenseCap outputs.
However, changing the semantics of subjects is less relevant than mentioning two different people in an image that presents only one, as seen in the Concat-Filter result.
In line 2, all methods generate wrong outputs for the image containing two people. A possible reason is the summarization model struggling with captions referring to different people in an image. In this case, we saw features from one person described as related to another. Nonetheless, other examples with two or more people in our evaluation were described coherently.

As limitations of our approach, we can list the restriction to a static image, the error propagation from the models used, the possibility of amplifying the error by standardizing the most frequent subject, mixing descriptions from people when the image contains more than one person, and the vanishing of descriptions during the summarization process. The latter can be observed in Fig.~\ref{fig:fig2} line 1, in which none of the characteristics age, emotion or scene were mentioned in the summary.

The lack of an annotated dataset for the problem of describing images for vision-impaired people impacts our experimental evaluation. 
When using the SIP dataset, even filtering the captions to keep only the phrases mentioning people, most of the Ground-Truth descriptions are generic.
Since our approach includes fine details about the speaker, such as age, emotion, and scene, Ours can present low scores, because the reference annotation does not contain such information.
In this case, we added relevant and extra information that negatively affects the metrics.

\section{\uppercase{Conclusions}}
\label{sec:conclusion}

In this paper, we proposed an approach for generating context of webinar images for visually-impaired people. Current methods are limited as assistive technologies since they tend to compress all the visual elements of an image into a single caption with rough details, or create multiple disconnected phrases to describe every single region of the image, which does not match the needs of vision-impaired people. 
Our approach combines Dense Captioning, as well as a set of filters to fit descriptions in our domain, with a language model for generating descriptive paragraphs focused on cognitive and visual details for visually impaired people. We evaluated and discussed with different metrics on the SIP dataset adapted to the problem. Experimentally, we demonstrated that a social inclusion method to improve the life of impaired people can be developed by combining existent methods for image analyze with neural linguistic models.   

As a future work, given the flexibility of our approach, we intend to replace and add more pretrained models to increase the accuracy and improve the quality of the generated context paragraphs. We also aim to create a new dataset designed to suit the image description problem for vision-impaired people.

\paragraph{\uppercase{Acknowledgements}.}
The authors thank CAPES, FAPEMIG and CNPq for funding different parts of this work, and the Google Cloud Platform for the awarded credits.

\bibliographystyle{apalike}
{\small
\bibliography{example}}

\end{document}